\documentclass[runningheads]{llncs}

\usepackage[utf8]{inputenc}
\usepackage[T1]{fontenc}
\usepackage[title]{appendix}
\usepackage{booktabs}   
\usepackage{graphicx}   
\usepackage[hidelinks]{hyperref}
\usepackage{multirow}  
\usepackage{xcolor}     
\usepackage{natbib}
\usepackage{float}
\usepackage{amsfonts}
\usepackage{todonotes}
\usepackage{subcaption}
\usepackage{amsmath}
\usepackage{url} 
\usepackage{algorithm}
\usepackage{algorithmic}
\usepackage{comment}
\usepackage{array}
\usepackage{enumitem}

\newcommand{\best}[1]{\textbf{#1}}

\title{Stein Variational Black-Box Combinatorial Optimization}
\titlerunning{Stein Variational Black-Box Combinatorial Optimization}

\author{Thomas Landais\and
Olivier Goudet \and
Adrien Goëffon \and
Frédéric Saubion \and
Sylvain Lamprier}

\authorrunning{T. Landais et al.}

\institute{LERIA, Université d’Angers, 2 Boulevard Lavoisier, Angers 49045, France\\
\email{\{thomas.landais,olivier.goudet,adrien.goeffon\}@univ-angers.fr}
\email{\{frederic.saubion,sylvain.lamprier\}@univ-angers.fr}
}

\date{February 2026}

\begin{document}

\maketitle

\begin{abstract}
Combinatorial black-box optimization in high-dimensional settings demands a careful trade-off between exploiting promising regions of the search space and preserving sufficient exploration to identify multiple optima. Although Estimation-of-Distribution Algorithms (EDAs) provide a powerful model‑based framework, they often concentrate on a single region of interest, which may result in premature convergence when facing complex or multimodal objective landscapes.
In this work, we incorporate the Stein operator to introduce a repulsive mechanism among particles in the parameter space, thereby encouraging the population to disperse and jointly explore several modes of the fitness landscape. Empirical evaluations across diverse benchmark problems show that the proposed method achieves performance competitive with, and in several cases superior to, leading state‑of‑the‑art approaches, particularly on large-scale instances. These findings highlight the potential of Stein variational gradient descent as a promising direction for addressing large, computationally expensive, discrete black-box optimization problems.
\end{abstract}

\section{Introduction}

Black-box optimization (BBO) deals with the challenge of optimizing an objective function whose internal structure or analytical form remains unknown. We focus here on the single-objective  optimization setting, where the goal is to maximize a function $f: \mathcal{X} \to \mathbb{R}$ over a discrete search space $\mathcal{X}$. In most practical scenarios, $f$ is computationally expensive to evaluate. Whether an evaluation involves running a high-fidelity simulation, conducting a physical experiment, or executing a complex algorithm, the high cost per query makes sample efficiency a key requirement \cite{audet2016blackbox, brochu2010tutorial}. 


 To solve BBO, two predominant families of approaches have emerged. On the one hand, Bayesian optimization \citep{forrester2009recent,frazier2018bayesian}  treats the unknown objective function $f$ as the outcome of a stochastic process, typically a Gaussian process.  On the other hand, Evolutionary Algorithms (EAs) evolve a population of candidate solutions through variation operators (mutation, crossover) and selection mechanisms based on fitness \cite{back1996evolutionary, eiben2015introduction}. Estimation-of-Distribution Algorithms (EDAs) are a subcategory of EAs \cite{larranaga2025estimation}. These stochastic methods guide the search for optima by explicitly learning and sampling from a probabilistic model of promising candidate solutions.

Despite their adaptive capabilities, EDAs often suffer from a premature collapse of the search distribution. Since the probabilistic model is  updated  iteratively based on the current population's performance, the search tends to concentrate rapidly around a single region of the search space, reducing diversity and potentially overlooking alternative optima. 

To address this limitation, particle-based methods have given rise to numerous approaches, often expressed through bio-inspired metaphors. While these metaphors have facilitated intuition, they have also drawn attention to the reuse of similar underlying mechanisms, which have been carefully examined \cite{AranhaCCDRSSS22,MOLINA2025102063}.

While these heuristics can be effective in practice, they generally lack the theoretical guarantees required for principled exploration of the fitness landscape. In this work, we propose to adapt the particle‑based dynamics of Stein Variational Gradient Descent (\texttt{SVGD}) \citep{liu2016stein} for black‑box optimization. \texttt{SVGD} transports a collection of particles toward a target distribution by combining functional‑gradient updates with kernel‑induced repulsion. This mechanism allows the algorithm to ascend efficiently toward high‑quality regions of the search space while preserving population diversity through a mathematically grounded repulsive term.

Recently, the integration of \texttt{SVGD} with \texttt{CMA‑ES} for continuous black‑box optimization has been investigated  \cite{braun2024stein}. In our work, we  demonstrate the feasibility of combining EDAs with \texttt{SVGD} for combinatorial black-box optimization, where applying direct gradient descent techniques is not possible. 

\vspace{0.5em}
Our contributions in this paper are as follows:

 \begin{itemize}[topsep=0pt]
\item We introduce a general framework that enables a principled extension of any EDA into a  multi‑EDA scheme that seek to approximate directly a target Boltzmann distribution.
\item  We propose a new \texttt{SVGD} objective that preserves invariance under monotonic transformations of the objective function. From this formulation, we derive a general rank‑based update rule suitable for any multi‑agent EDAs.
\item   We provide a proof of concept of our framework that extends univariate EDAs in a multi-agent EDA with \texttt{SVGD}. Empirical evidence indicates that repulsion arising from the kernel plays a role in maintaining population diversity, facilitating exploration in large discrete spaces where traditional EDAs commonly experience premature convergence. These early results suggest promising potential for converging toward high‑quality solutions in challenging combinatorial landscapes.
  \end{itemize}
%


The remainder of this paper is organized as follows. First, we describe the background and position our work (Section~2). We then introduce our \texttt{SVGD-EDA} algorithm (Section~3). Next, we provide an empirical evaluation through a series of experiments on benchmark tasks (Section~4). Finally, we discuss future research directions in the conclusion.

\section{Background and Motivations 
\label{sec:problem_setting}}


We consider a discrete search space $\mathcal{X}=\mathcal{X}_1\times\cdots\times\mathcal{X}_n$ and aim to solve the maximization problem $\max_{x\in\mathcal{X}} f(x)$, where $x=(x_1,\dots,x_n)$ denotes a candidate solution in $\mathcal{X}$. 


In this section, we begin by summarizing the core principles of Estimation-of-Distribution Algorithms (EDAs), before explaining why we shift toward a broader perspective centered on approximating a target distribution.
We then present the Stein Variational Gradient Descent method \citep{liu2016stein}, which we will use in the next section to leverage our probabilistic formulation of combinatorial optimization as a variational inference problem, free from the constraints imposed by a restricted parametric family of distributions.

\subsection{From Estimation-of-Distribution Algorithms (EDAs) to Variational Inference \label{sec:EDA_var}}

EDAs are population-based algorithms that frame optimization as an iterative density estimation process, where the objective is to learn a probability distribution that increasingly concentrates its mass on high-quality regions of the search space. Instead of relying on genetic operators such as crossover or mutation, an EDA evolves a distribution $\pi_{\theta}$ from a parametric family $\Pi$, with parameters 
$\theta \in \Theta$. 

Thus, rather than directly dealing with the former optimization problem over $\mathcal{X}$, EDAs consider an equivalent reformulation of the search  over the parameter space $\Theta$ of a probabilistic model $\pi_\theta$: $ \max_{\theta \in \Theta}\; \mathbb{E}_{x\sim \pi_\theta}\big[f(x)\big]$.  
At each iteration $t$, the algorithm performs three steps:  
(i) a population of $\lambda$ samples $x^{1},\dots,x^{\lambda}$ is generated from the current model $\pi_{\theta^t}$;  
(ii) the objective function is evaluated on these samples;  
(iii) the parameter vector is updated to $\theta^{t+1}$ so as to increase the likelihood of generating high-fitness solutions. This cycle continues until a predefined stopping criterion is reached.

From an information-theoretic viewpoint, EDAs can be embedded in a more general
variational framework in which we consider a target Boltzmann distribution over the
parameter space $\Theta$, expressed as:
\begin{equation}
p(\theta) \propto \exp\!\left(\frac{1}{\gamma}\,\mathbb{E}_{x\sim\pi_\theta}[f(x)]\right).
\label{eq:boltzmann_target}
\end{equation}
Following the maximum-entropy principle, the distribution $q\in\mathcal{Q}$ that
minimizes the Kullback–Leibler divergence from $p$ is
\begin{equation}
q^*=\arg\min_{q\in\mathcal{Q}} D_{\mathrm{KL}}(q\|p)
= \arg\max_{q\in\mathcal{Q}}
\mathbb{E}_{\theta\sim q}\mathbb{E}_{x\sim\pi_\theta}[f(x)]
+ \gamma \mathcal{H}(q),
\label{kl}
\end{equation}
where $\mathcal{H}(q)$ denotes the entropy of $q$. This formulation highlights
the role of the temperature parameter $\gamma$: as $\gamma\to 0$, $p$ becomes
sharply peaked, and the optimal $q$ collapses to a Dirac distribution centered
at the maximizer $\theta^*$. In this regime, corresponding to classical EDAs, exploration over $\mathcal{X}$ arises solely from the epistemic uncertainty
encoded in the generator $\pi_\theta$ and the diversity of high-quality sampled
solutions. 
Because traditional EDAs typically rely on unimodal generator
models, such as multivariate Gaussians (\texttt{CMA‑ES}) \citep{hansen2001completely} or factorized Bernoulli models
(\texttt{PBIL}) \citep{baluja1994population}, the resulting search is constrained to represent only simple, single-mode
structures, often leading to premature convergence on multimodal landscapes

We argue that optimizing \eqref{kl} with $\gamma>0$ can mitigate such collapse
by explicitly regularizing the search distribution through its entropy, thereby
preserving diversity at the level of $q(\theta)$, even when the underlying
generator $\pi_\theta$ remains unimodal. 
Under
this perspective, the variational inference problem acts as a landscape
exploration mechanism: it aims to discover and characterize the dominant basins
of attraction that structure the fitness landscape, enabling subsequent
intensification within these basins to obtain high-quality solutions.

 
To address this variational inference problem without restricting the search to a
predefined parametric family $\mathcal{Q}$, we propose to rely on Stein
Variational Gradient Descent (\texttt{SVGD}). The next subsection introduces \texttt{SVGD} in its
original formulation, while Section~\ref{method} adapts it to our combinatorial
optimization setting, leveraging a set of interacting particles to capture
complex, potentially multimodal distributions over $\Theta$.

\subsection{Stein Variational Gradient Descent (\texttt{SVGD}) \label{sec:svgd}}

Given a target distribution $p$ on a space $\Theta$, the \texttt{SVGD} algorithm \cite{liu2016stein} considers a tractable family $\mathcal{Q}$ of probability measures on $\Theta$, where each $q\in\mathcal{Q}$ is represented by the empirical distribution of particles $\{\theta_i\}_{i=1}^m$. The method updates an initial empirical measure $q_0$ through transport maps
\[
T_t(\theta)=\theta+\epsilon_t\,\phi_{q_t,p}^\star(\theta), 
\qquad q_{t+1}=q_{t[T_t]},
\]
so as to decrease $\mathrm{KL}(q_t\|p)$.

The vector field $\phi_{q_t,p}^\star$ is given by the kernelized Stein operator
\[
\phi_{q_t,p}^\star(\theta)
=\mathbb{E}_{\theta'\sim q_t}
\!\left[
k(\theta',\theta)\,\nabla_{\theta'} \log p(\theta')
+ \nabla_{\theta'} k(\theta',\theta)
\right].
\]
In practice, with particles (or samples) $\{\theta_i^t\}_{i=1}^m\sim q_t$, the updates become $\theta_i^{t+1}
\leftarrow \theta_i^t + \epsilon_t\,\widehat{\phi}^\star(\theta_i^t)$ with 
\begin{equation}
\widehat{\phi}^\star(\theta_i^t)
=\frac{1}{m}\sum_{j=1}^m
\left[
k(\theta_j^t,\theta_i^t)\,\nabla_{\theta_j}\log p(\theta_j)\Bigr\rvert_{\theta_j = \theta^t_j}
+ \nabla_{\theta_j}k(\theta_j,\theta_i^t)\Bigr\rvert_{\theta_j = \theta^t_j}
\right].
\label{eq:svgd_empirical}
\end{equation}

\begin{figure}[htbp]
    \centering
    \begin{subfigure}[b]{0.48\linewidth}
        \centering
        \includegraphics[width=\linewidth]{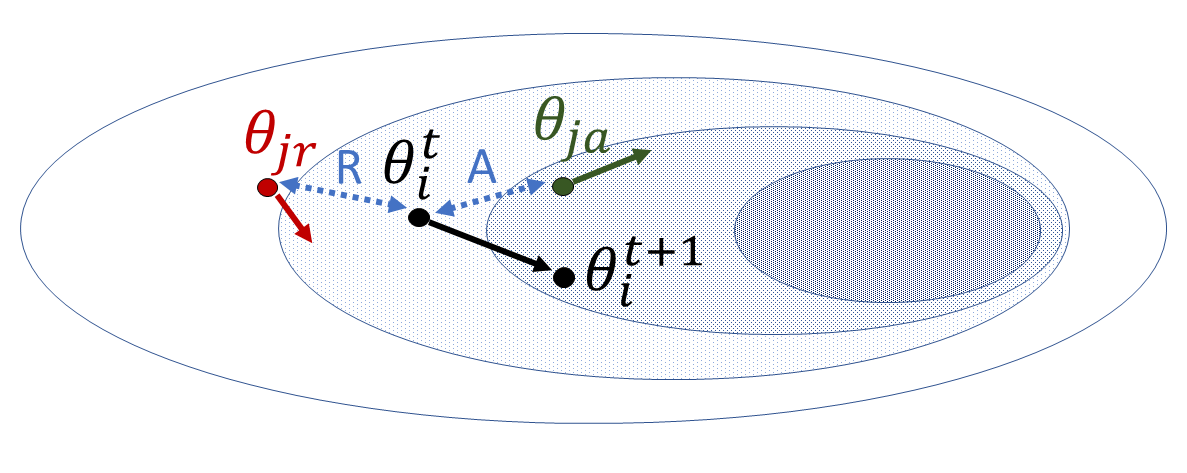}
        \caption{\texttt{SVGD} particle update from $\theta^t_i$ to $\theta^{t+1}_i$, using \eqref{eq:svgd_empirical}. Green particle $\theta_{ja}$ induces attraction via $k(\theta_{ja},\theta^t_i)$; red particle $\theta_{jr}$ induces repulsion.}
        \label{fig:svgd_update}
    \end{subfigure}
    \hfill
    \begin{subfigure}[b]{0.48\linewidth}
        \centering
        \includegraphics[width=\linewidth]{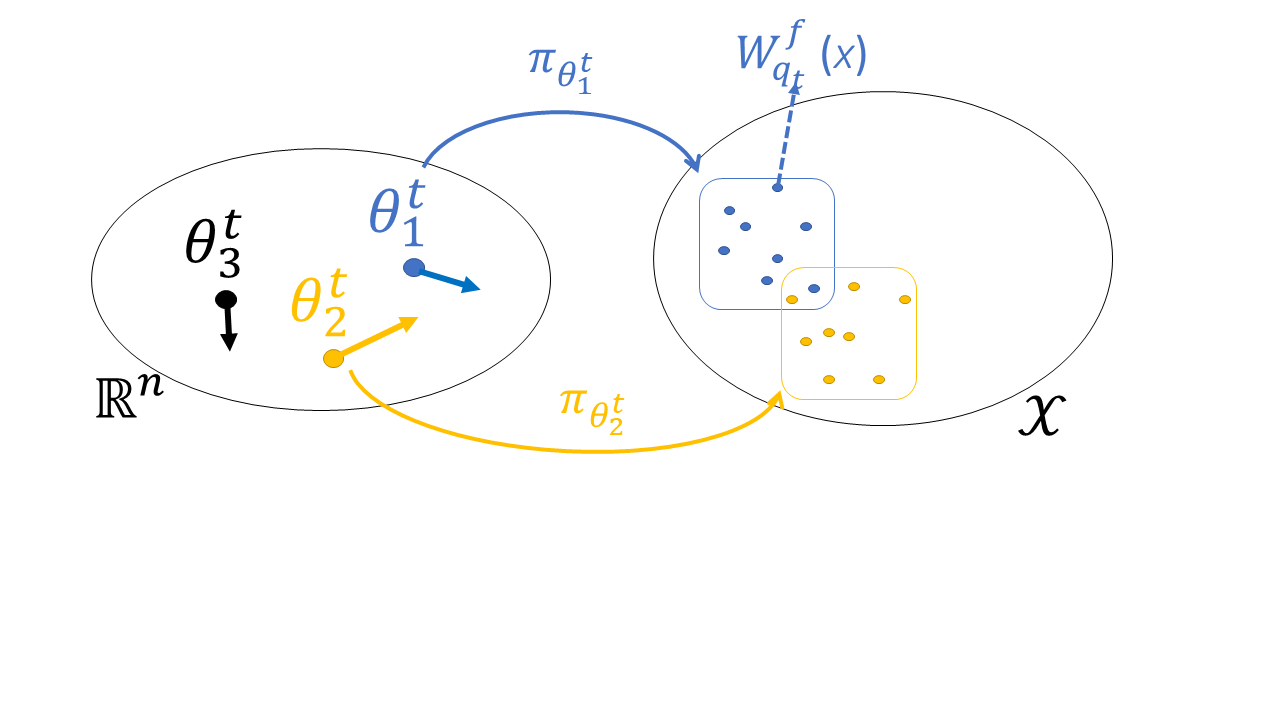}
        \caption{\texttt{SVGD-EDA} process. At step $t$, particles $\theta^t_i$ sample solutions in ${\cal X}$ via $\pi_{\theta^t_i}$. Solutions estimate $W^f_{q_t}(x)$ (Section \ref{sec:invariant_svgd}).}
        \label{fig:svgd_eda}
    \end{subfigure}
    \caption{Illustration of \texttt{SVGD} mechanisms:  
    (a)~\texttt{SVGD-EDA} at particle‑level,  
    (b)~\texttt{SVGD-EDA} process, including solution sampling.}
    \label{fig:svgd_mechanisms}
\end{figure}

The kernel term involving $\nabla_{\theta_j}k()$ induces repulsion, whereas the term weighted by $\nabla_{\theta_j}\log p()$ transports particles toward regions of high target density (See Fig. \ref{fig:svgd_update}). These interactions prevent collapse and allow SVGD to capture multimodal targets, providing a flexible nonparametric alternative to parametric inference. 




\section{Stein variational combinatorial optimization}
\label{method}


This section begins by presenting the proposed methodology, which leverages the \texttt{SVGD} framework to directly approximate the target Boltzmann distribution. We then introduce a variant featuring a modified objective function that remains invariant under any monotonic transformation of the original objective. Finally, we derive practical update rules that enable a multi‑agent EDA to tackle discrete black‑box optimization problems. In this formulation, each agent’s parameter vector is interpreted as a particle within the \texttt{SVGD} algorithm.

\subsection{Derivation of a multi-agent Estimation of distribution algorithm with \texttt{SVGD}}
\label{sec:multi-agentSVGD}

We formalize a multi-agent EDA as a collection of $m$ parameter vectors $\{\theta_i\}_{i=1}^m$, where each  $\theta_i$ parameterizes a policy acting as the underlying probabilistic model of an agent. At each iteration~$t$, every agent $i$ generates $\lambda$ solutions. Within the \texttt{SVGD} framework presented in Section \ref{sec:svgd}, each parameter vector $\theta^t_i$ is treated as a sample from a time-varying empirical distribution $q_t$. Our goal is to make $q$ evolve so that it converges toward the target Boltzmann distribution $p(\theta)$, defined by \eqref{eq:boltzmann_target}.

First, leveraging the log-derivative trick  on the Boltzmann target \eqref{eq:boltzmann_target}, for $j=1, \dots, m$, we estimate  $\nabla_{\theta_j} \log p(\theta_j)\Bigr\rvert_{\theta_j = \theta^t_j}$  using the $\lambda$ solutions $\{x^{j(\ell)}\}_{\ell=1}^\lambda$ generated by agent $j$:

\begin{align}
\nabla_{\theta_j} \log p(\theta_j)\Bigr\rvert_{\theta_j = \theta^t_j} &= \frac{1}{\gamma}  \int_{x \in \mathcal{X}}f(x) \nabla_{\theta_j} \text{log}\  \pi_{\theta_j}(x)\Bigr\rvert_{\theta_j = \theta^t_j} \pi_{\theta^t_j}(x)dx \\
&\approx  \frac{1}{\lambda\gamma} \sum_{\ell=1}^{\lambda} f(x^{j(\ell)}) \nabla_{\theta_j} \text{log}\  \pi_{\theta_j}(x^{j(\ell)})\Bigr\rvert_{\theta_j = \theta^t_j}.
\end{align}

Then, by substituting this approxmation into the empirical Stein operator \eqref{eq:svgd_empirical},  we obtain the following  update rule for each agent $i$ at iteration $t$, which is influenced by all parameter vectors $\{\theta^t_1, \dots, \theta^t_m\}$ through
\begin{equation}
\scriptsize
\theta_i^{t+1} \leftarrow \theta_i^t + \epsilon  \frac{1}{m} \sum_{j=1}^m \left[ \frac{k(\theta_i^t, \theta_j^t)}{\lambda\gamma} \sum_{\ell=1}^{\lambda} f(x^{j(\ell)}) \nabla_{\theta_j} \log \pi_{\theta_j}(x^{j(\ell)})\Bigr\rvert_{\theta_j = \theta^t_j} + \nabla_{\theta_j} k(\theta_i^t, \theta_j)\Bigr\rvert_{\theta_j = \theta^t_j} \right].
\label{eq:update_rule_f}
\end{equation}

It gives us a first version of multi-agent EDA with \texttt{SVGD} allowing 
the maximization of $\mathbb{E}_{x \sim \pi_\theta}[f(x)]$ through the variational approximation of the target distribution \eqref{eq:boltzmann_target}.  

However, this update rule is highly sensitive to extreme values of the objective function $f$. As a consequence, the parameters involved in \eqref{eq:update_rule_f}, such as the learning rate $\epsilon$, must be carefully tuned to the scale and variability of the problem‑specific fitness landscape. By contrast, a desirable property exhibited by many widely used EDA algorithms, such as \texttt{CMA-ES} \citep{hansen2001completely} and \texttt{PBIL} \citep{baluja1994population}, is the invariance of the search dynamics under monotonic transformations of the objective function. This ensures that algorithmic behavior does not depend on the absolute scaling of $f$, thereby improving robustness across heterogeneous optimization tasks. This motivates the development, presented in the next section, of an alternative formulation whose objective inherently preserves such invariance.

\subsection{Invariance‑preserving objective for black-box optimization with \texttt{SVGD} \label{sec:invariant_svgd}}

Drawing on the Information‑Geometric Optimization (IGO) framework \citep{ollivier2017information}, the raw objective $f$ can be replaced by a monotone, quantile‑based transformation $W_{\theta^t}^f$ that depends on the current parameter vector $\theta^t$ of the generator. In \citep{ollivier2017information}, the optimization problem  $\max_{\theta \in \Theta}\; \mathbb{E}_{x\sim \pi_\theta}\big[f(x)\big]$ (see Section \ref{sec:EDA_var}) 
is accordingly reformulated at each iteration $t$ as $\max_{\theta \in \Theta}\; \mathbb{E}_{x \sim \pi_\theta}\!\left[ W_{\theta^t}^f(x) \right]$, with $W_{\theta^t}^f(x) := w\!\left( \mathbb{P}_{x' \sim \pi_{\theta^t}}\!\left(f(x') > f(x)\right) \right)$ where $w$ is a non‑increasing function.

However, in our context, since we aim to evolve a distribution $q$ over parameters $\theta$ with \texttt{SVGD}, the objective cannot depend on a single $\theta^t$. Instead, we propose to optimize $J^f_t(\theta) =  \mathbb{E}_{x \sim \pi_\theta}\big[W_{q_t}^f(x)\big]$, with $W_{q_t}^f(x)=w\bigl(
\mathbb{E}_{\theta^t \sim q_t} \Bigl[\mathbb{P}_{x' \sim \pi_{\theta^t}}\bigl(f(x') > f(x)\bigr)\Bigr]\bigr)$ and $w: [0,1] \rightarrow \mathbb{R} $ a non-increasing function. 

At iteration $t$, this objective can be interpreted as searching for parameters $\theta$ that maximize, up to the monotone transformation $w$, the probability that samples drawn from $\pi_{\theta}$ outperform in average those generated by the previous generation of EDA agents $\pi_{\theta^t}$ with  $\theta^t \sim q_t$. 

This transformation promotes consistent policy improvement throughout the search process, and guarantees invariance with respect to any monotonic transformation of the objective function $f$. 

The Boltzmann distribution that we seek to approximate becomes then at iteration $t$:  $p_t(\theta) \propto \exp\!\left( \frac{1}{\gamma} 
\mathbb{E}_{x \sim \pi_\theta}\big[W_{q_t}^f(x)\big]\right)$.

When searching to approximate this target distribution with \texttt{SVGD}, the update direction for each agent at each iteration $t$ becomes then 
\begin{equation}
\scriptsize 
\theta_i^{t+1} \leftarrow \theta_i^t + \epsilon  \frac{1}{m} \sum_{j=1}^m \left[ \frac{k(\theta_i^t, \theta_j^t)}{\lambda\gamma} \sum_{\ell=1}^{\lambda} W_{q_t}^f(x^{j(\ell)}) \nabla_{\theta_j} \log \pi_{\theta_j}(x^{j(\ell)})\Bigr\rvert_{\theta_j = \theta^t_j} + \nabla_{\theta_j} k(\theta_i^t, \theta_j)\Bigr\rvert_{\theta_j = \theta^t_j} \right], 
\label{eq:final_update_rule}
\end{equation}

\noindent where an unbiased estimator of $W_{q_t}^f(x^{j(\ell)})$ using sampled solutions at iteration t is given by \\
\small
$w \left(\frac{1}{m}(\sum_{\substack{i=1, i\neq j}}^{m}
 \frac{1}{\lambda}\big|\{k: f(x^{i(k)})>f(x^{j(\ell)})\}\big| +  \frac{1}{\lambda-1} \big|\{k\neq \ell:f(x^{j(k)})>f(x^{j(\ell)})\}\big|) \right).$



\normalsize
The choice of the weighting function $w$ used to compute the score $W_{q_t}^f(x^{j(\ell)})$ for each candidate solution plays a central role in shaping the selection pressure exerted during the update of agent parameters. By transforming the ordinal ranks of individuals into utility values, the function $w$ determines the relative contribution of each sampled point when updating the model. This principle is fundamental to ranked-based optimization methods such as \texttt{CMA‑ES}, where careful tuning of utility weights has been shown to substantially influence performance across diverse classes of optimization problems \citep{andersson2015parameter}.

In this work, we adopt a simplified form of utility weighting by introducing the function $w : [0,1] \to \mathbb{R}$ defined as $w(x) = 1 - 2x$. When this function is used and all sampled points $x^{j(\ell)}$ are distinct,  $W_{q_t}^f(x^{j(\ell)})$ can be approximated by $\hat{S}(x^{j(\ell)}) = 1 - 2 \,\frac{\operatorname{rk}(x^{j(\ell)}, \Gamma^t, f)}{m\lambda - 1}$, where $\Gamma^t = \{x^{i(k)}\}_{i = 1..m,\; k = 1..\lambda}$ denotes the complete population at iteration $t$, and
$\operatorname{rk}(x, \Gamma, f) = \big|\{x' \in \Gamma : f(x') > f(x)\}\big|$ is the number of individuals in $\Gamma$ with strictly higher fitness (with ties broken at random). Thus, under this very simple scoring scheme, the best solution of the complete population sampled at iteration $t$ attains rank $0$ and obtains score $+1$, while the worst receives rank $m\lambda - 1$ and score~$-1$.

\subsection{Practical implementation for black-box  discrete optimization}

In this section, we implement the general framework of the multi-agent EDA presented in the previous section to solve discrete black-box  optimization problems (See Fig. \ref{fig:svgd_eda}). This algorithm is called \texttt{SVGD-EDA}. The version of \texttt{SVGD-EDA} designed to solve pseudo-Boolean problems is summarized in Algorithm \ref{alg:svgd_eda}.

For pseudo-Boolean problems, the probabilistic model associated with each agent is chosen to be a univariate  Bernoulli distribution.\footnote{More expressive multivariate discrete probabilistic models, such as those employed in MIMIC \citep{de1996mimic},  BOA \citep{pelikan2002bayesian}, or more recently in RL-EDA \cite{goudet2025black} could also be incorporated to capture dependencies between variables, but exploring such extensions is left for future work.}
In this case, for each agent $i \in [1..m]$,  the parameter vector $\theta_i = (\theta_{i1}, \dots, \theta_{in}) \in \mathbb{R}^n$ specifies the log-odds of the binary variables. The probability of sampling a solution $x \in \mathcal{X}$ from $\pi_{\theta_i}$ is 
$\pi_{\theta_i} (x) = \prod_{l=1}^n \sigma(\theta_{il})^{x_l} (1 - \sigma(\theta_{il}))^{1 - x_l}$, with $\sigma(z) = (1 + e^{-z})^{-1}$ the sigmoid function. 

When the search space is categorical, $\mathcal{X} = [1..D]^n$. For each agent $i$ and variable $l$, we maintain a vector of logits $\theta_{il} = (\theta_{il1}, \dots, \theta_{ilD}) \in \mathbb{R}^D$. The softmax function converts these into the categorical distribution $\mathbb{P}(x_l = d \mid \theta_{il}) = \frac{\exp(\theta_{ild})}{\sum_{d'=1}^D \exp(\theta_{ild'})}$. Assuming independence across variables, the probability of sampling a solution $x \in \mathcal{X}$ from the distribution $\pi_{\theta_i}$ is given by $\pi_{\theta_i}(x) = \prod_{l=1}^n \frac{\exp(\theta_{il,x_k})}{\sum_{d=1}^D \exp(\theta_{ild})}.$

Regarding the choice of the kernel function $k$, we propose to employ the standard Radial Basis Function (RBF) defined as   $k(\theta_i, \theta_j) = \exp\!\left( - \frac{ \|\theta_i - \theta_j\|^2}{2h^2} \right)$, where $h > 0$ is the bandwidth. To ensure that the kernel scale remains appropriate throughout optimization, the bandwidth $h$ is adapted using the median trick of \cite{liu2016stein}.

\vspace{-1em}
 \begin{algorithm}[!h]
 \caption{\texttt{SVGD-EDA} for pseudo-Boolean problems with parameters $m, \lambda \in \mathbb{N}^*$, $\epsilon, \gamma  \in \mathbb{R}$}
 \label{alg:svgd_eda}
  \begin{algorithmic}[1]
  \STATE \textbf{Input:} instance $(\mathcal{X},f)$, with $\mathcal{X} = \{0,1\}^n$, objective $f : \mathcal{X} \rightarrow \mathbb{R}$ and number of iterations $T$.
  
  \STATE \# Initialization
  \FOR{$i = 1$ \TO $m$}
      \STATE Randomly initialize the parameter vector 
             $\theta_i^0 = (\theta_{i1}^0,\dots,\theta_{in}^0)$
  \ENDFOR
  \STATE $f^* \leftarrow -\infty$
  
  \FOR{$t = 0$ \TO $T-1$}
     \STATE \# Generation and evaluation of solutions 
    \FOR{$j = 1$ \TO $m$}
        \FOR{$\ell = 1$ \TO $\lambda$}
            \STATE $x^{j(\ell)} \sim \pi_{\theta^t_j}$
            \STATE $f' \leftarrow f(x^{j(\ell)})$
            \IF{ $f' > f^*$}
                \STATE $x^* \leftarrow x^{j(\ell)}$
            \ENDIF
        \ENDFOR
    \ENDFOR
    \STATE \# Update EDAs parameters using \texttt{SVGD}
    \FOR{$i = 1$ \TO $m$}
        \STATE $\theta_i^{t+1} \leftarrow \theta_i^t + \epsilon  \frac{1}{m} \sum_{j=1}^m \left[ \frac{k(\theta_i^t, \theta_j^t)}{\lambda\gamma} \sum_{\ell=1}^{\lambda} \hat{S}(x^{j(\ell)}) \nabla_{\theta_j} \log \pi_{\theta_j}(x^{j(\ell)}) +  \nabla_{\theta_j} k(\theta_i^t, \theta_j)\right]$
    \ENDFOR   
 \ENDFOR  
  \STATE \textbf{Output:} best solution $x^*$
  \end{algorithmic}
\end{algorithm}

\vspace{-2em}
\section{Experiments}

The objectives of this section are twofold. First, we benchmark the performance of the proposed multi-agent EDA against established state-of-the-art methods for pseudo-Boolean black-box optimization. Second, we conduct an ablation study to identify the critical algorithmic components that facilitate effective exploration of the search space, providing insight into the mechanism underlying the algorithm’s diversity-preserving properties.

The source code, the instances and the complete results for all algorithms are available at the \url{https://github.com/LandOS973/Supplementary-material}.



\subsection{Experimental protocol and parameters setting}
\label{sec:protocol}


We focus on the NKD model which is a natural extension of the NK model of Kauffman \citep{kauffman1989nk} to cases where variables can take more than two categorical values. This is a framework for describing fitness landscapes whose problem size and ruggedness are both parameterizable. The NKD function is defined as $f_{\rm NKD}(x) = \frac{1} {n} \sum_{i=1}^n \gamma_i(x_i,x_{l_{i1}},\dots,x_{l_{iK}})$, with each subfunction $\gamma_i:\{0,1, \dots, D_i-1\}^{K+1}\rightarrow [0,1[$ is defined over categorical variables with $D_i$ possible values for each variable.  We generate instances with $D_i=2$ for every variable, which corresponds to the original pseudo-boolean NK problem, but we also generate instances of a categorical problem called NK3 with $D_i=3$ for every variable. 

For each variant NK or NK3 of the problem four different types of distribution of instances with $K \in \{1,2,4,8\}$ are built. Increasing $K$ expands the size of each epistatic neighborhood, which in turn increases the ruggedness of the landscape and typically leads to a larger number of local optima. For each combinatorial problem, we conducted evaluations across three problem sizes $n \in \{64, 128, 256\}$ and we generated 10 distinct instances for each configuration $(n,K)$. Following a black-box optimization setting, we assume no prior knowledge of the objective function or of the interaction structure between variables.

The behavior of the multi-agent \texttt{SVGD-EDA} depends on several key hyperparameters. Across all experiments, we use a fixed configuration with $m=7$, $\lambda=13$, and $\gamma=0.015$, selected via a grid search over a representative set of problem instances. This choice is not critical: the performance remains stable and qualitatively similar for a range of nearby parameter settings.

Thanks to a fully tensorial PyTorch implementation, \texttt{SVGD-EDA} efficiently exploits hardware acceleration for large-scale evaluations. Empirical measurements on an Intel Core i5-12400F CPU and an NVIDIA RTX 3060 GPU show that solving a single nk instance ($n=256$) with a budget of 50,000 iterations takes respectively 7 seconds and 3 seconds (including fitness evaluations). By leveraging GPU vectorization, an entire batch of 100 independent runs with 50,000 iterations is processed simultaneously in about 20 seconds on the GPU, compared to 300 seconds on the CPU for this size of instance.


\subsection{Experimental Validation on Discrete Black-Box Benchmarks}
\label{sec:benchmark_validation}

We evaluate the performance of our multi-agent \texttt{SVGD-EDA} against a comprehensive set of 80 baseline algorithms from the \texttt{Nevergrad} library (v1.0.12) \cite{nevergrad}, which includes classic metaheuristics (Evolutionary Strategies, Differential Evolution) and machine-learning-driven adaptive portfolios, often represented by multiple variants of the same algorithm with different hyperparameter configurations. We enriched this pool with three classical  EDAs: \texttt{PBIL} \cite{baluja1994population}, \texttt{MIMIC} \cite{de1996mimic}, and \texttt{BOA} \cite{boa}, using their default hyperparameter settings and source code available at \url{https://github.com/e5120/EDAs}. This yields a total of 84 evaluated algorithms (our method and 83 baselines) for the pseudo-Boolean NK instances. However, because \texttt{PBIL} is strictly limited to binary search spaces, it was excluded from the experiments on the categorical NK3 instances, reducing the total ranking pool to 83 algorithms for that specific problem class. We impose a budget of 50,000 objective function evaluations per run, consistent with recent evolutionary algorithm benchmarking protocols with this number of functions evaluations for this sizes of instances  \citep{ye2020benchmarking,ye2022automated}. To ensure statistical robustness, each algorithm is evaluated over 100 independent runs per $(n,K)$ problem configuration (10 runs on 10 distinct instances).

\subsubsection{Global experimental results}
Table \ref{tab:results_portrait} presents a selection of these results, comparing \texttt{SVGD-EDA} to three other EDAs of the same category: \texttt{PBIL} \cite{baluja1994population}, \texttt{MIMIC} \cite{de1996mimic}, and \texttt{BOA} \cite{boa}. The final columns report the performance of the best algorithm among all remaining competitors, including the 80 Nevergrad discrete algorithms \cite{nevergrad}. Based on this average score, the algorithms are ranked, and their position among all competitors is indicated. 


\begin{table}[htbp]
    \centering
    \resizebox{\textwidth}{!}{%
    \begin{tabular}{ccc cc cc cc cc lcc}
        \toprule
        \multicolumn{3}{c}{\textbf{Instances}} & 
        \multicolumn{2}{c}{\textbf{\texttt{SVGD-EDA}}} & 
        \multicolumn{2}{c}{\textbf{\texttt{PBIL}}} & 
        \multicolumn{2}{c}{\textbf{\texttt{MIMIC}}} & 
        \multicolumn{2}{c}{\textbf{\texttt{BOA}}} & 
        \multicolumn{3}{c}{\textbf{Best Method (Others)}} \\
        \cmidrule(r){1-3} \cmidrule(lr){4-5} \cmidrule(lr){6-7} \cmidrule(lr){8-9} \cmidrule(lr){10-11} \cmidrule(l){12-14}
        Pb & $n$ & $t$ & Rank & Score & Rank & Score & Rank & Score & Rank & Score & Name & Rank & Score \\
        \midrule
        \multirow{12}{*}{NK} & \multirow{4}{*}{64} & 1 & 2/84 & 0.7124 & 36/84 & 0.7097 & 54/84 & 0.7052 & 56/84 & 0.7008 & \mbox{\texttt{DiscreteDE}} & \best{1/84} & \best{0.7126} \\
         &  & 2 & 2/84 & 0.7475 & 10/84 & 0.7412 & 53/84 & 0.7312 & 56/84 & 0.7262 & \mbox{\texttt{DiscreteDE}} & \best{1/84} & \best{0.7497} \\
         &  & 4 & 2/84 & 0.7609 & 4/84 & 0.7492 & 54/84 & 0.7330 & 55/84 & 0.7316 & \mbox{\texttt{DiscreteDE}} & \best{1/84} & \best{0.7658} \\
         &  & 8 & \best{1/84} & \best{0.7538} & 8/84 & 0.7368 & 54/84 & 0.7176 & 55/84 & 0.7174 & \mbox{\texttt{DiscreteDE}} & 2/84 & 0.7535 \\
        \cmidrule(lr){2-14}
         & \multirow{4}{*}{128} & 1 & \best{1/84} & \best{0.7118$^{*}$} & 3/84 & 0.7085 & 54/84 & 0.6956 & 56/84 & 0.6928 & \mbox{\texttt{DiscreteDE}} & 2/84 & 0.7094 \\
         &  & 2 & \best{1/84} & \best{0.7406$^{*}$} & 3/84 & 0.7334 & 55/84 & 0.7147 & 56/84 & 0.7135 & \mbox{\texttt{DiscreteDE}} & 2/84 & 0.7364 \\
         &  & 4 & \best{1/84} & \best{0.7646$^{*}$} & 3/84 & 0.7514 & 54/84 & 0.7229 & 53/84 & 0.7285 & \mbox{\texttt{DiscreteDE}} & 2/84 & 0.7535 \\
         &  & 8 & \best{1/84} & \best{0.7454$^{*}$} & 3/84 & 0.7369 & 55/84 & 0.7034 & 56/84 & 0.6681 & \mbox{\texttt{DiscreteDE}} & 2/84 & 0.7373 \\
        \cmidrule(lr){2-14}
         & \multirow{4}{*}{256} & 1 & \best{1/84} & \best{0.7091$^{*}$} & 2/84 & 0.7048 & 56/84 & 0.6802 & 50/84 & 0.6863 & \mbox{\texttt{DiscreteDE}} & 3/84 & 0.7029 \\
         &  & 2 & \best{1/84} & \best{0.7395$^{*}$} & 2/84 & 0.7314 & 54/84 & 0.6989 & 48/84 & 0.7093 & \mbox{\texttt{HugeLognormalDiscreteOnePlusOne}} & 3/84 & 0.7269 \\
         &  & 4 & \best{1/84} & \best{0.7598$^{*}$} & 2/84 & 0.7429 & 55/84 & 0.7019 & 47/84 & 0.7162 & \mbox{\texttt{HugeLognormalDiscreteOnePlusOne}} & 3/84 & 0.7352 \\
         &  & 8 & \best{1/84} & \best{0.7334$^{*}$} & 2/84 & 0.7315 & 55/84 & 0.6805 & 63/84 & 0.5852 & \mbox{\texttt{OnePtRecombiningDiscreteLenglerOnePlusOne}} & 3/84 & 0.7242 \\
        \midrule[1.2pt]
        \multirow{12}{*}{NK3} & \multirow{4}{*}{64} & 1 & 2/83 & 0.7894 & — & — & 54/83 & 0.7659 & 55/83 & 0.7651 & \mbox{\texttt{DiscreteDE}} & \best{1/83} & \best{0.7918} \\
         &  & 2 & \best{1/83} & \best{0.8204$^{*}$} & — & — & 51/83 & 0.7825 & 54/83 & 0.7793 & \mbox{\texttt{DiscreteDE}} & 2/83 & 0.8154 \\
         &  & 4 & \best{1/83} & \best{0.8066$^{*}$} & — & — & 55/83 & 0.7645 & 58/83 & 0.7548 & \mbox{\texttt{DiscreteDE}} & 2/83 & 0.8033 \\
         &  & 8 & 14/83 & 0.7627 & — & — & 59/83 & 0.6583 & 57/83 & 0.7000 & \mbox{\texttt{DiscreteDE}} & \best{1/83} & \best{0.7779$^{*}$} \\
        \cmidrule(lr){2-14}
         & \multirow{4}{*}{128} & 1 & \best{1/83} & \best{0.7975$^{*}$} & — & — & 48/83 & 0.7600 & 50/83 & 0.7530 & \mbox{\texttt{DiscreteDE}} & 2/83 & 0.7918 \\
         &  & 2 & \best{1/83} & \best{0.8123$^{*}$} & — & — & 48/83 & 0.7621 & 55/83 & 0.7523 & \mbox{\texttt{DiscreteDE}} & 2/83 & 0.7959 \\
         &  & 4 & \best{1/83} & \best{0.7989$^{*}$} & — & — & 54/83 & 0.7469 & 56/83 & 0.7321 & \mbox{\texttt{RandRecombiningDiscreteLenglerOnePlusOne}} & 2/83 & 0.7829 \\
         &  & 8 & 44/83 & 0.7491 & — & — & 63/83 & 0.6098 & 58/83 & 0.6297 & \mbox{\texttt{OnePtRecombiningDiscreteLenglerOnePlusOne}} & \best{1/83} & \best{0.7618$^{*}$} \\
        \cmidrule(lr){2-14}
         & \multirow{4}{*}{256} & 1 & \best{1/83} & \best{0.7889$^{*}$} & — & — & 47/83 & 0.7363 & 49/83 & 0.7238 & \mbox{\texttt{DiscreteLengler3OnePlusOne}} & 2/83 & 0.7741 \\
         &  & 2 & \best{1/83} & \best{0.8035$^{*}$} & — & — & 47/83 & 0.7393 & 49/83 & 0.7249 & \mbox{\texttt{DiscreteLengler3OnePlusOne}} & 2/83 & 0.7839 \\
         &  & 4 & \best{1/83} & \best{0.7823$^{*}$} & — & — & 47/83 & 0.7265 & 54/83 & 0.7072 & \mbox{\texttt{DiscreteLengler3OnePlusOne}} & 2/83 & 0.7783 \\
         &  & 8 & 46/83 & 0.7285 & — & — & 63/83 & 0.5784 & 62/83 & 0.5790 & \mbox{\texttt{SmallLognormalDiscreteOnePlusOne}} & \best{1/83} & \best{0.7552$^{*}$} \\
        \midrule
        \multicolumn{3}{c}{\textbf{SVGD-EDA}} & 5.375 & 0.763 & \multicolumn{2}{c}{} & \multicolumn{2}{c}{} & \multicolumn{2}{c}{} & \multicolumn{1}{r}{\textbf{DiscreteDE}} & 10.625 & 0.756 \\
        \bottomrule
    \end{tabular}%
    }
    \caption{Global rankings and average scores at 50,000 evaluations (100 runs). \textbf{Bold} values indicate the best mean score. An asterisk ($^*$) denotes a statistically significant improvement over the second-best method (Wilcoxon signed-rank test on instance-wise means, $p < 0.05$). The final row summarizes the global mean ranks and mean scores across all instances.}
    \label{tab:results_portrait}
\end{table}

As seen in Table \ref{tab:results_portrait}, 
\texttt{SVGD-EDA}  reaches the 1st rank in 17 out of the 24 tested distributions of instances  (and obtain the best average  rank of 5.375 across $(n,K)$ configurations). When not ranked 1st, the algorithm is very frequently 2nd. \texttt{SVGD-EDA}  consistently shows superior performance compared to other estimation of distribution algorithms such as \texttt{PBIL}, \texttt{MIMIC}, or \texttt{BOA}. On pseudo-boolean NK landscapes, it demonstrates strong scalability as the instance size increases with $n=128$ and $n=256$, where it systematically leads the 84 algorithms across all ruggedness levels ($K=1, 2, 4, 8$). On NK3 instances, \texttt{SVGD-EDA} dominates the rankings for $n=128$ and $n=256$, with the exception of the most constrained cases ($K=8$) where specific  \texttt{OnePlusOne} variants maintain an edge. 

This trend suggests that the cooperative mechanism driven by the \texttt{SVGD} kernel is effective in high-dimensional spaces. By maintaining population diversity through agent interaction, the method appears to mitigate the premature convergence and stagnation that affect standard independent agents or simpler EDAs in larger search spaces. 


\subsubsection{Evolution of the average scores during the search}

\begin{figure}[h]
    \centering
    \begin{subfigure}[b]{0.54\textwidth}
        \centering
\includegraphics[width=\linewidth]{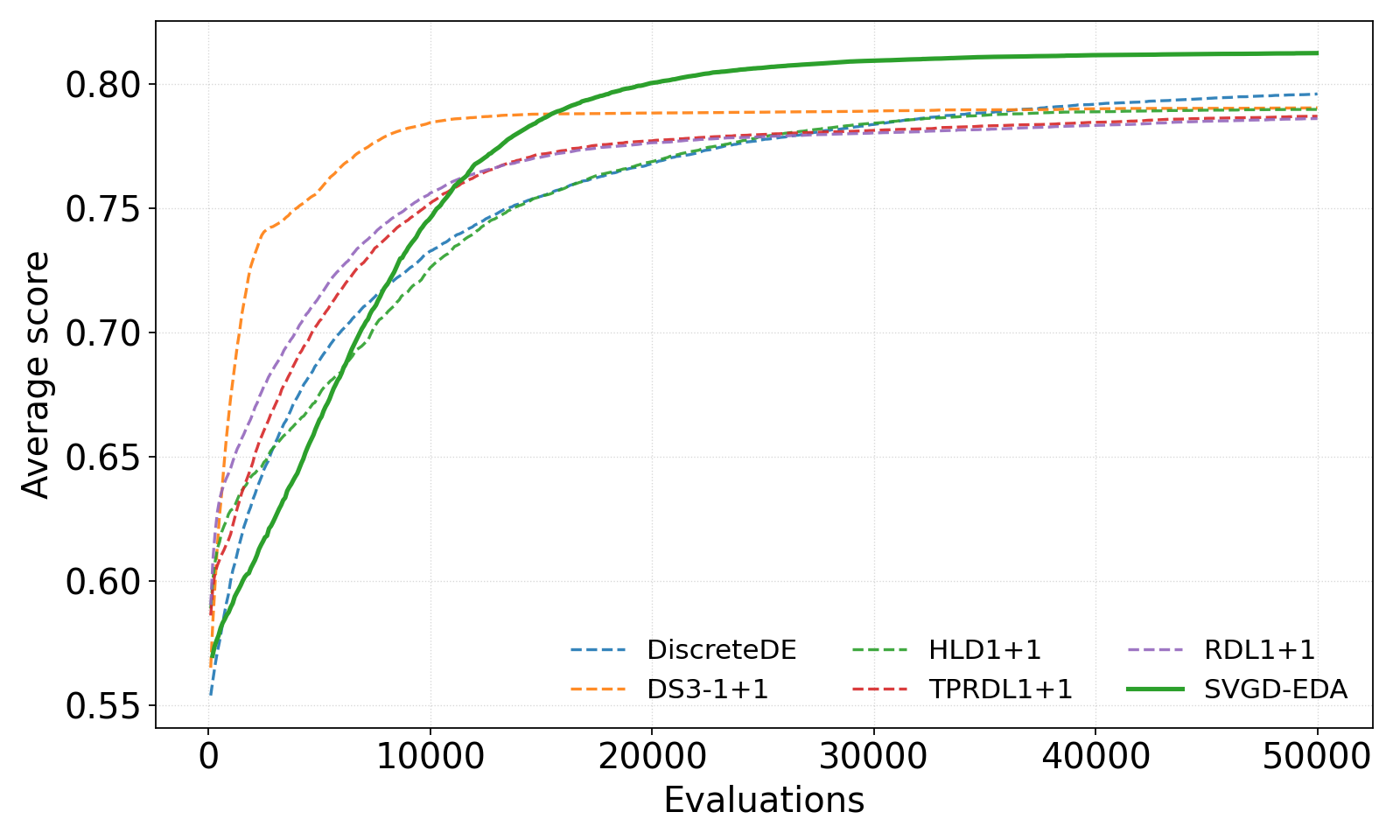}        

    \end{subfigure}
    \begin{subfigure}[b]{0.44\textwidth}
    \centering
\includegraphics[width=\linewidth]{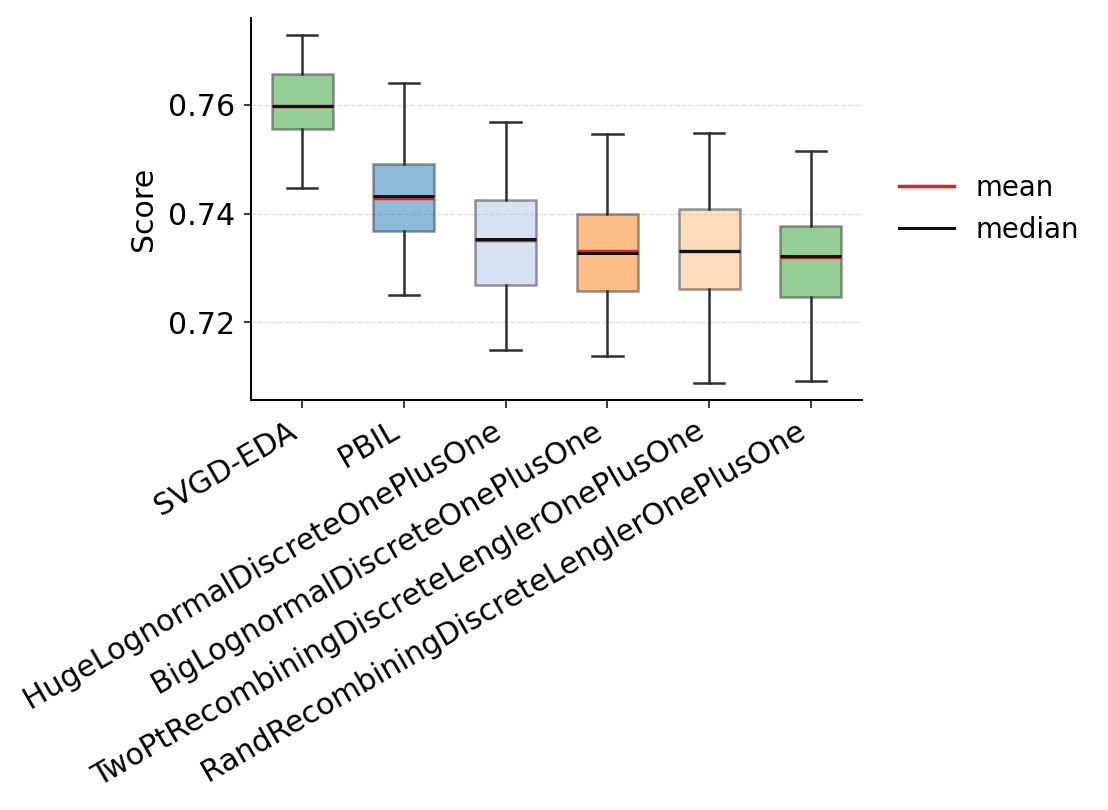}

    \end{subfigure}
    \caption{Performance analysis of \texttt{SVGD-EDA} with top-five competitors on an NK3 instance distribution ($n = 128, K = 2$). 
    Left:   Evolution of the average score. Right: Final score distribution after 50,000 evaluations. 
    Acronyms in legends:
    \texttt{\scriptsize DS3-1+1}~$\rightarrow$~\texttt{\scriptsize DiscreteLengler3OnePlusOne},  
    \texttt{\scriptsize HLD1+1}~$\rightarrow$~\texttt{\scriptsize HugeLognormalDiscreteOnePlusOne}, 
    \texttt{\scriptsize TPRDL1+1}~$\rightarrow$~\texttt{\scriptsize TwoPtRecombiningDiscreteLenglerOnePlusOne}, \\
    \texttt{\scriptsize RDL1+1}~$\rightarrow$~\texttt{\scriptsize RecombiningDiscreteLenglerOnePlusOne}.
    }

\label{fig:benchmark_plots_nk3_128_t2}
\end{figure}

Figure \ref{fig:benchmark_plots_nk3_128_t2} on the left illustrates the evolution of the average best score from the start of the search on a categorical NK3 landscape ($n=128$, $K=2$, with three categories per variable).
Green curve corresponds to  \texttt{SVGD-EDA}  compared to the five other best-performing competitors (over 83) with  the budget of 50,000 evaluations. From this plot, we observe that \texttt{DiscreteLengler3OnePlusOne} (orange curve), an elitist (1+1) evolutionary strategy with an decreasing mutation scope, exhibits faster initial progress than \texttt{SVGD-EDA}. 
Its favorable performance for small optimization budgets stems from its large initial mutation neighborhood, which allow for fast progress by combining multiple elementary moves~\cite{lengler2015fixed,doerr2019self}. 
However, the progressive reduction of this neighborhood turns the algorithm into a 1-flip random hill-climber, leading to early stagnation once a local optimum is reached.
In contrast, \texttt{SVGD-EDA} exhibits sustained improvement throughout the optimization process and ultimately attains better final results. Figure \ref{fig:benchmark_plots_nk3_128_t2} on the right compares the distribution of final scores with those of the five best competing algorithms. \texttt{SVGD-EDA} not only attains a higher mean performance but also exhibits strong reliability, as evidenced by a tight interquartile range across 100 independent runs, indicating consistent solution quality.

    

\subsection{Sensitivity Analysis: Impact of the Number of EDA agents $m$}
\label{sec:ablation}

To assess the influence of the multi-agent architecture and the repulsive Stein force on optimization performance, we conducted a sensitivity analysis on the number of EDA agents, denoted by $m$.

We evaluated the framework across different sizes of instances ($n \in \{64, 128,$ $256\}$) on NK and NK3. We tested 16 configurations for the number of agents: $m \in [1..16]$. The other hyperparameters stay at their baseline value as described in Section \ref{sec:protocol}. These experiments were conducted under a strict fixed budget of $50,000$ evaluations. Consequently, an increase in $m$ mechanically reduces the total number of generations available for the EDA agents to converge, as the budget per EDA agent is approximately $50,000 / m$. This setup tests the trade-off between broad parallel exploration and deep local exploitation. 
Finally, to ensure statistical robustness, each reported result represents the aggregate of 100 independent runs.

\begin{figure}[h]
     \centering
     \begin{subfigure}[b]{0.48\textwidth}
         \centering
         \includegraphics[width=\textwidth]{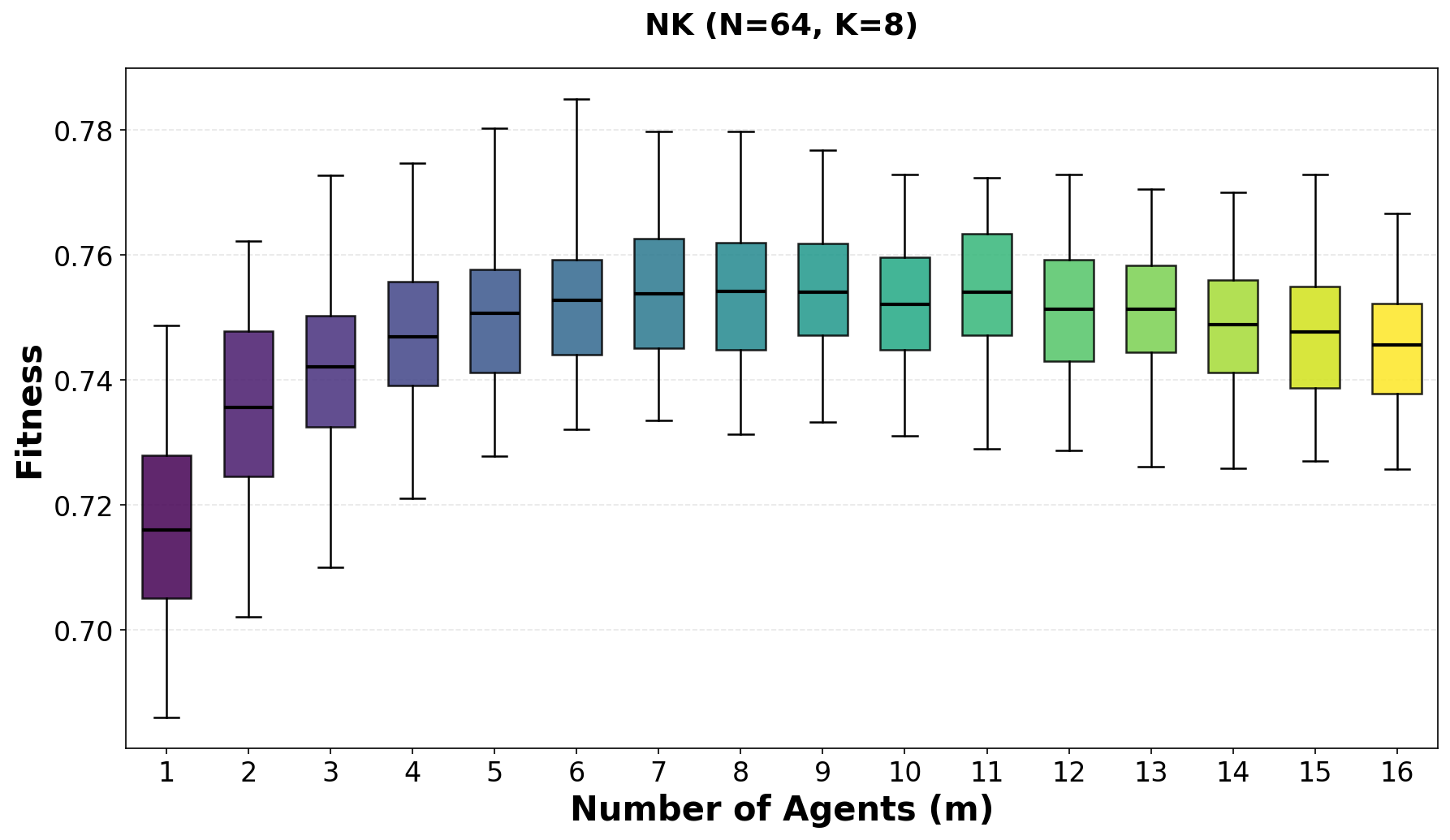}
         \caption{NK $n=64, K=8$}
     \end{subfigure}
     \hfill
     \begin{subfigure}[b]{0.48\textwidth}
         \centering
         \includegraphics[width=\textwidth]{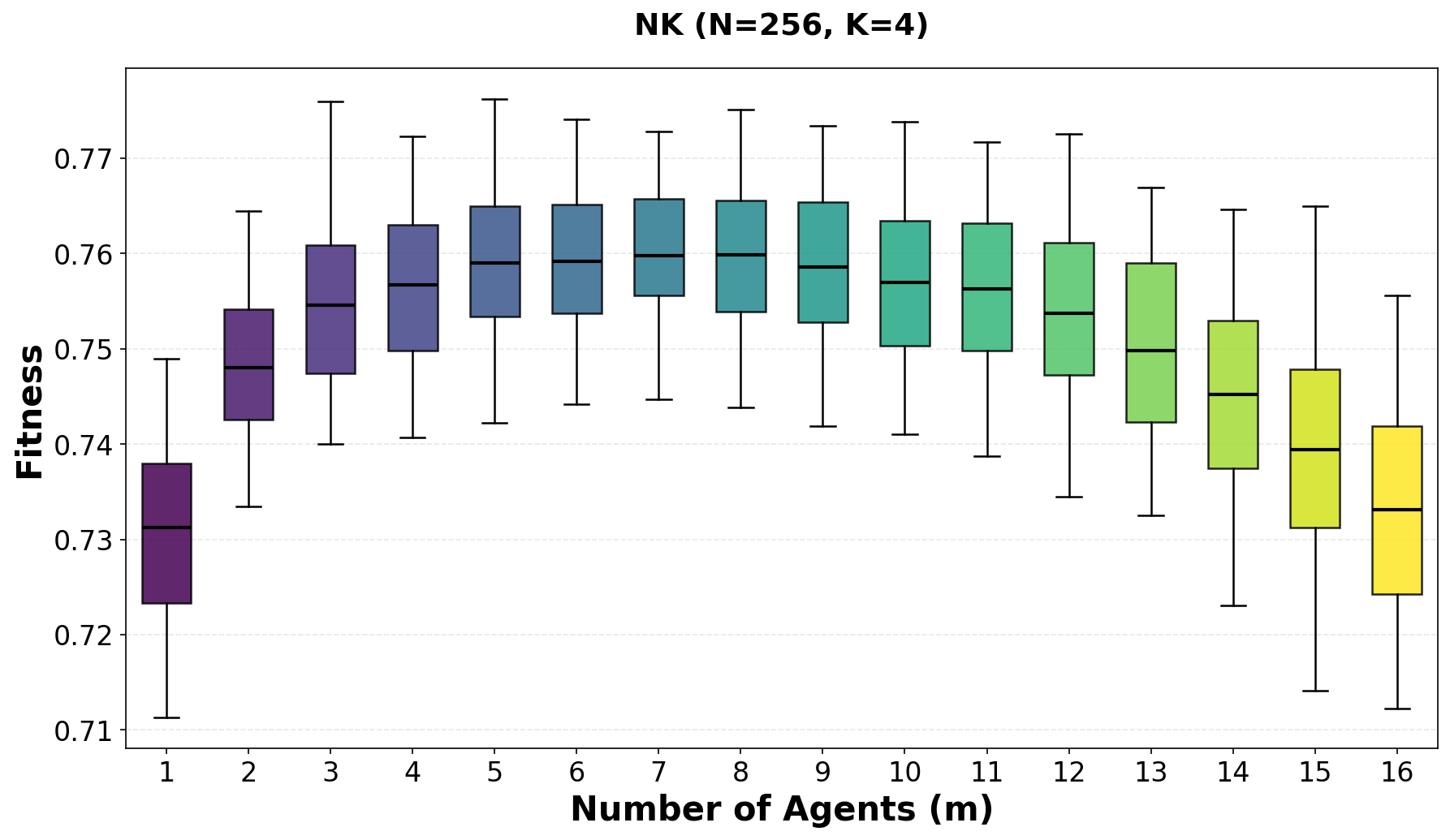}
         \caption{NK $n=256, K=4$}
     \end{subfigure}
     \caption{Sensitivity analysis on NK landscapes. The boxplots illustrate the distribution of final fitness scores across 100 independent runs for varying numbers of EDA agents $m$ (x-axis). Higher scores indicate better performance.}
     \label{fig:sensi_nk}
\end{figure}

Figure \ref{fig:sensi_nk} displays the distribution of the final fitness scores achieved at the end of the 50,000 evaluations for different EDA agent counts $m \in [1..16]$. These box-plots reveal a distinct behavioral shift governed by the interaction between dimensionality and ruggedness.

On moderate-dimensional problems ($n=64$), deploying a larger number of EDA agents ($m \in [10..14]$) is generally beneficial to maximize broad exploration. However, as the dimensionality increases to $n=256$ or as the landscape becomes more rugged (e.g., $K=8$), the optimal configuration systematically shifts towards a more compact EDA agent count ($m \approx 5$ to $7$). This phenomenon suggests that the optimal $m$ is not solely a function of dimensionality, but rather a consequence of the combined complexity: both higher dimensionality and increased ruggedness  amplify the need for a larger individual evaluation budget per EDA agent to ensure convergence.

Importantly, performance degrades noticeably across all problem types for $m > 14$. This is directly attributed to \textit{budget dilution}: with too many EDA agents under a strictly fixed evaluation budget, the number of generations available per EDA becomes insufficient to perform the necessary model refinements. In conclusion, for expensive black-box optimization, a balanced configuration (such as $m=7$) appears to offer the most robust compromise between collective landscape coverage and individual search depth.

\subsection{Ablation study on the impact of Stein attraction/repulsion mechanisms in \texttt{SVGD-EDA}}
\label{sec:ablation_interaction}

To isolate the specific contribution of the Stein repulsive force, we conducted an ablation study comparing the standard multi-agent \texttt{SVGD-EDA} ($m=7$) against a baseline of non-interacting EDA agents. In the ``No Interaction'' setting, the RBF kernel in \eqref{eq:final_update_rule} is replaced by the identity matrix (i.e., $k(\theta_i, \theta_j) = \delta_{ij}$) and the gradient of the kernel term $\nabla_{\theta_j} k(\theta_i, \theta_j)$ is forced to zero. This effectively decouples the EDA agents, reducing the algorithm to $m$ independent EDAs  running in parallel.

\begin{figure}[htbp]
    \centering
    \includegraphics[width=0.75\textwidth]{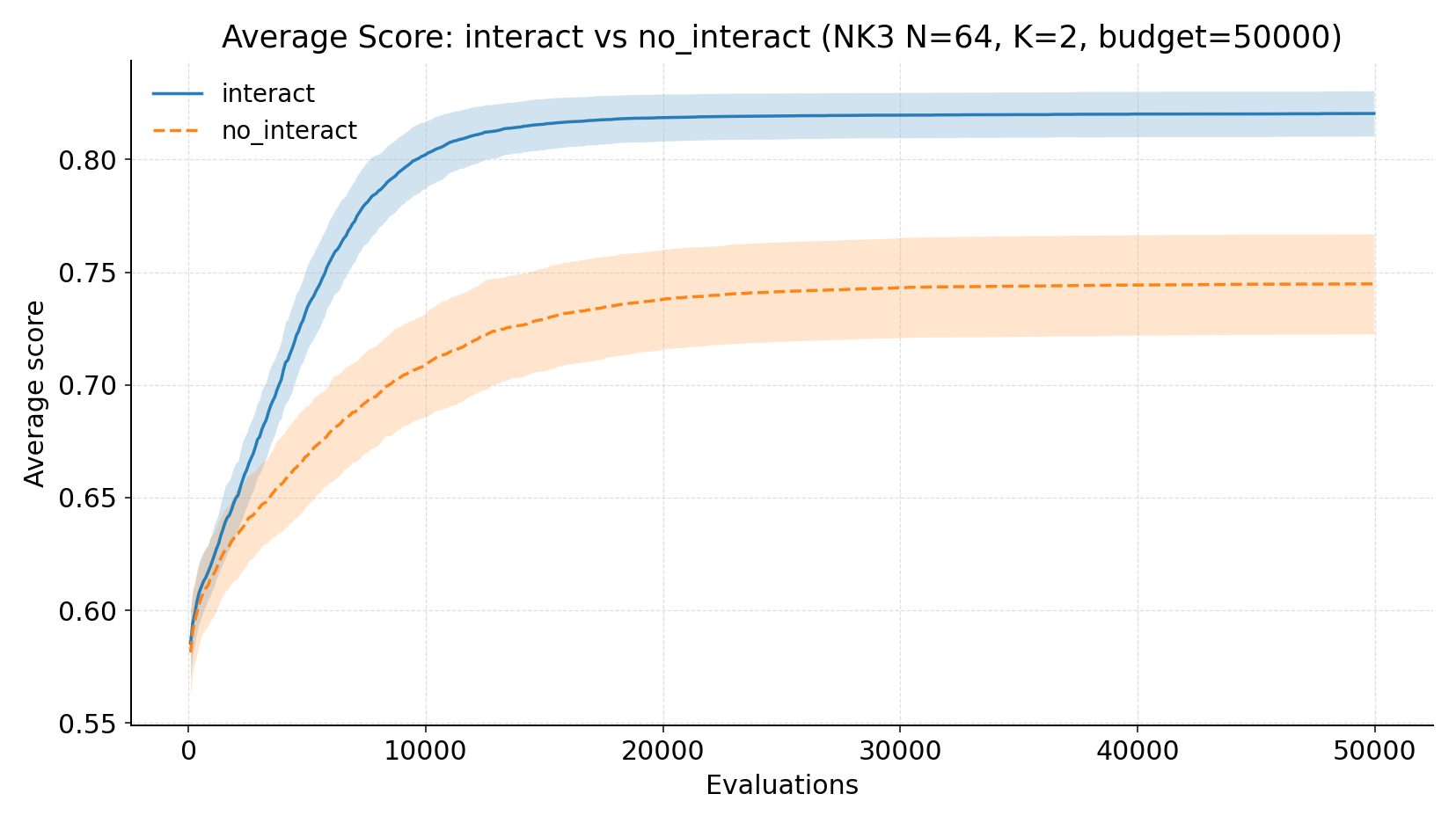}
    \caption{Impact of Interaction on NK3 with $n=64$ and  $K=2$. Comparison of average score evolution between cooperative (blue) and independent (orange) agent EDAs. The repulsive force prevents early stagnation in deceptive local optima.}
    \label{fig:interact_vs_no_nk3}
\end{figure}

While detailed tabular results over all configurations are omitted for brevity, aggregated data reveal a clear correlation between problem complexity and the utility of cooperation. On simple, low-dimensional binary spaces (e.g., NK $n=64, K \le 4$), independent agents perform similarly to the interacting ones. However, as the dimensionality increases, the landscape becomes more rugged, or the search space shifts to a categorical domain (NK3), the interacting approach systematically outperforms the independent baseline. For instance, performance gaps reach up to +15\,\% on large, highly rugged NK landscapes ($n=256, K=8$) and consistently exceed +10\,\% across all categorical NK3 instances.

To illustrate the mechanics behind this performance gap, Figure \ref{fig:interact_vs_no_nk3} highlights a distinct behavioral advantage provided by the multi-agent interaction on a specific NK3 instance ($n=64, K=2$). Although both methods initially progress and appear to plateau around 20,000 evaluations, the independent baseline gets permanently trapped in a sub-optimal region. In contrast, the kernel-induced repulsion maintains enough population diversity for the interacting EDA agents to eventually escape this deceptive local optimum and locate a superior global solution ($\approx +10\,\%$ performance gap).

\section{Conclusion}

In this paper, we introduced a principled multi-agent Estimation-of-Distribution Algorithm (\texttt{SVGD-EDA}) designed for complex, high-dimensional black-box combinatorial optimization. By explicitly mapping the Stein Variational Gradient Descent framework to the parameter space of independent Bernoulli and categorical distributions, our approach structurally prevents the premature variance collapse that typically afflicts standard EDAs and local search heuristics. 

Our empirical validation on large-scale instances ($n=256$) of binary NK and categorical NK3 landscapes demonstrates that \texttt{SVGD-EDA} effectively avoids early stagnation and sustains continuous improvement throughout the search. The ablation study further isolates and quantifies the contribution of  kernel-induced repulsion, revealing performance improvements of up to 15\,\% over non-interacting parallel EDAs by enabling the population to bypass deceptive local optima. 


Future work will focus on two concrete technical extensions. First, given the limitations of the Euclidean RBF kernel applied to logit parameters, we will investigate information-geometric kernels (e.g., Fisher-Rao or Jensen-Shannon) defined directly on the statistical manifold, so that repulsive forces reflect meaningful distributional divergences. Second, to mitigate the budget dilution effect while preserving the exploratory benefits of a large population, we plan to study sparse or asynchronous updates schemes, in which only a subset of agents is updated at each iteration.

 \begin{credits}

\subsubsection{\discintname}
The authors have no competing interests to declare that are relevant to the content of this article.

\end{credits}

\bibliographystyle{plain}
\bibliography{references}

@incollection{larranaga2025estimation,
  title={Estimation of distribution algorithms},
  author={Larranaga, Pedro and Bielza, Concha},
  booktitle={Handbook of Heuristics},
  pages={583--598},
  year={2025},
  publisher={Springer}
}

@article{AranhaCCDRSSS22,
  author       = {Claus Aranha and
                  Christian Leonardo Camacho{-}Villal{\'{o}}n and
                  Felipe Campelo and
                  Marco Dorigo and
                  Rub{\'{e}}n Ruiz and
                  Marc Sevaux and
                  Kenneth S{\"{o}}rensen and
                  Thomas St{\"{u}}tzle},
  title        = {Metaphor-based metaheuristics, a call for action: the elephant in
                  the room},
  journal      = {Swarm Intell.},
  volume       = {16},
  number       = {1},
  pages        = {1--6},
  year         = {2022},
  url          = {https://doi.org/10.1007/s11721-021-00202-9},
  doi          = {10.1007/S11721-021-00202-9},
  bibsource    = {dblp computer science bibliography, https://dblp.org}
}

@article{MOLINA2025102063,
title = {The paradox of success in evolutionary and bioinspired optimization: Revisiting critical issues, key studies, and methodological pathways},
journal = {Swarm and Evolutionary Computation},
volume = {98},
pages = {102063},
year = {2025},
issn = {2210-6502},
doi = {https://doi.org/10.1016/j.swevo.2025.102063},
url = {https://www.sciencedirect.com/science/article/pii/S2210650225002214},
author = {Daniel Molina and Javier {Del Ser} and Javier Poyatos and Francisco Herrera}
}

@article{de1996mimic,
  title={MIMIC: Finding optima by estimating probability densities},
  author={De Bonet, Jeremy and Isbell, Charles and Viola, Paul},
  journal={Advances in neural information processing systems},
  volume={9},
  year={1996}
}

@article{NK,
  title={The NK model of rugged fitness landscapes and its application to maturation of the immune response},
  author={Kauffman, Stuart A and Weinberger, Edward D},
  journal={Journal of theoretical biology},
  volume={141},
  number={2},
  pages={211--245},
  year={1989},
  publisher={Elsevier}
}

@techreport{baluja1994population,
  title={Population-based incremental learning. a method for integrating genetic search based function optimization and competitive learning},
  author={Baluja, Shumeet},
  year={1994}
}

@incollection{frazier2018bayesian,
  title={Bayesian optimization},
  author={Frazier, Peter I},
  booktitle={Recent advances in optimization and modeling of contemporary problems},
  pages={255--278},
  year={2018},
  publisher={Informs}
}

@article{forrester2009recent,
  title={Recent advances in surrogate-based optimization},
  author={Forrester, Alexander IJ and Keane, Andy J},
  journal={Progress in aerospace sciences},
  volume={45},
  number={1-3},
  pages={50--79},
  year={2009},
  publisher={Elsevier}
}

@inproceedings{andersson2015parameter,
  title={Parameter tuned {CMA-ES} on the {CEC'15} expensive problems},
  author={Andersson, Martin and Bandaru, Sunith and Ng, Amos HC and Syberfeldt, Anna},
  booktitle={2015 IEEE congress on evolutionary computation (CEC)},
  pages={1950--1957},
  year={2015},
  organization={IEEE}
}

@book{back1996evolutionary,
  title={Evolutionary algorithms in theory and practice: evolution strategies, evolutionary programming, genetic algorithms},
  author={Back, Thomas},
  year={1996},
  publisher={Oxford university press}
}

@article{kauffman1989nk,
  title={The {NK} model of rugged fitness landscapes and its application to maturation of the immune response},
  author={Kauffman, Stuart A and Weinberger, Edward D},
  journal={Journal of theoretical biology},
  volume={141},
  number={2},
  pages={211--245},
  year={1989},
  publisher={Elsevier}
}

@book{eiben2015introduction,
  title={Introduction to evolutionary computing},
  author={Eiben, Agoston E and Smith, James E},
  year={2015},
  publisher={Springer}
}

@article{boa,
author = {Pelikan, Martin and Goldberg, D.E. and Cantu-Paz, Erick},
year = {1999},
month = {01},
pages = {525-532},
title = {BOA: The {B}ayesian optimization algorithm},
journal = {BOA: The Bayesian Optimization Algorithm}
}

@misc{nevergrad,
    author = {J. Rapin and O. Teytaud},
    title = {{Nevergrad - A gradient-free optimization platform}},
    year = {2018},
    publisher = {GitHub},
    journal = {GitHub repository},
    howpublished = {\url{https://GitHub.com/FacebookResearch/Nevergrad}},
}

@book{pelikan2002bayesian,
  title={Bayesian optimization algorithm: From single level to hierarchy},
  author={Pelikan, Martin},
  year={2002},
  publisher={University of Illinois at Urbana-Champaign}
}

@article{hansen2001completely,
  title={Completely derandomized self-adaptation in evolution strategies},
  author={Hansen, Nikolaus and Ostermeier, Andreas},
  journal={Evolutionary Computation},
  volume={9},
  number={2},
  pages={159--195},
  year={2001}
}

@article{goudet2025black,
  title={Black-Box Combinatorial Optimization with Order-Invariant Reinforcement Learning},
  author={Goudet, Olivier and Suire, Quentin and Go{\"e}ffon, Adrien and Saubion, Fr{\'e}d{\'e}ric and Lamprier, Sylvain},
  journal={arXiv preprint arXiv:2510.01824},
  year={2025}
}

@article{ollivier2017information,
  title={Information-geometric optimization algorithms: A unifying picture via invariance principles},
  author={Ollivier, Yann and Arnold, L{\'e}onard and Auger, Anne and Hansen, Nikolaus},
  journal={Journal of Machine Learning Research},
  volume={18},
  number={18},
  pages={1--65},
  year={2017}
}

@article{audet2016blackbox,
  title={Blackbox and derivative-free optimization: theory, algorithms and applications},
  author={Audet, Charles and Kokkolaras, Michael},
  journal={Optimization and Engineering},
  volume={17},
  number={1},
  pages={1--2},
  year={2016},
  publisher={Springer}
}

@article{brochu2010tutorial,
  title={A tutorial on Bayesian optimization of expensive cost functions, with application to active user modeling and hierarchical reinforcement learning},
  author={Brochu, Eric and Cora, Vlad M and de Freitas, Nando},
  journal={arXiv preprint arXiv:1012.2599},
  year={2010}
}

@article{liu2016stein,
  title={Stein variational gradient descent: A general purpose bayesian inference algorithm},
  author={Liu, Qiang and Wang, Dilin},
  journal={Advances in neural information processing systems},
  volume={29},
  year={2016}
}

@inproceedings{ye2020benchmarking,
  title={Benchmarking a genetic algorithm with configurable crossover probability},
  author={Ye, Furong and Wang, Hao and Doerr, Carola and B{\"a}ck, Thomas},
  booktitle={International Conference on Parallel Problem Solving from Nature},
  pages={699--713},
  year={2020},
  organization={Springer}
}

@article{ye2022automated,
  title={Automated configuration of genetic algorithms by tuning for anytime performance},
  author={Ye, Furong and Doerr, Carola and Wang, Hao and B{\"a}ck, Thomas},
  journal={IEEE Transactions on Evolutionary Computation},
  volume={26},
  number={6},
  pages={1526--1538},
  year={2022},
  publisher={IEEE}
}

@article{braun2024stein,
  title={Stein Variational Evolution Strategies},
  author={Braun, Cornelius V and Lange, Robert T and Toussaint, Marc},
  journal={arXiv preprint arXiv:2410.10390},
  year={2024}
}

@inproceedings{lengler2015fixed,
  title={Fixed budget performance of the (1+1) {EA} on linear functions},
  author={Lengler, Johannes and Spooner, Nicholas},
  booktitle={Proceedings of the 2015 ACM Conference on Foundations of Genetic Algorithms XIII},
  pages={52--61},
  year={2015}
}

@inproceedings{doerr2019self,
  title={Self-adjusting mutation rates with provably optimal success rules},
  author={Doerr, Benjamin and Doerr, Carola and Lengler, Johannes},
  booktitle={Proceedings of the Genetic and Evolutionary Computation Conference},
  pages={1479--1487},
  year={2019}
}

\appendix

\end{document}